\documentclass{article} 
\usepackage{iclr2016_conference,times}
\usepackage{hyperref}
\usepackage{url}
\usepackage{graphicx}
\usepackage{xspace}
\usepackage{caption}
\usepackage{amsmath} 
\usepackage{amssymb}
\DeclareMathOperator*{\argmax}{arg\,max}
\DeclareMathOperator*{\softmax}{softmax}
\usepackage{color}
\usepackage{enumitem}
\usepackage{placeins}

\newcommand{\starspace}{{\sc Embedding Model}\xspace}
\newcommand{\window}{{\sc window}\xspace}

\newcommand{\bA}{{\bf A}}
\newcommand{\bB}{{\bf B}}

\newcommand{\bH}{{\bf H}}
\newcommand{\bU}{{\bf U}}
\newcommand{\bm}{{\bf m}}
\newcommand{\bc}{{\bf c}}
\newcommand{\bq}{{\bf q}}
\newcommand{\ba}{{\bf a}}
\renewcommand{\Re}{\mathbb{R}}

\title{The Goldilocks Principle: Reading Children's Books with Explicit Memory Representations}

\author{Felix Hill\thanks{The majority of this work was done while FH was at Facebook AI Research, and was completed at his current affiliation, University of Cambridge, Computer Laboratory, Cambridge, UK.},~ Antoine Bordes, Sumit Chopra \& Jason Weston\\
Facebook AI Research\\
770 Broadway\\
New York, USA\\
\texttt{felix.hill@cl.cam.ac.uk,\{abordes,spchopra,jase\}@fb.com}
}

\iclrfinalcopy 

\begin{document}

\maketitle

\begin{abstract}
  We introduce a new test of how well language models capture meaning
  in children's books. Unlike standard language modelling benchmarks,
  it distinguishes the task of predicting syntactic function words
  from that of predicting lower-frequency words, which carry greater
  semantic content.  We compare a range of
  state-of-the-art models, each with a different way of
  encoding 
  what has been previously read. 
  We show that models which store explicit representations of long-term
  contexts outperform state-of-the-art neural language models at
  predicting semantic content words, although this advantage is not
  observed for syntactic function words.
 Interestingly, we find that the amount of
  text encoded in a single memory representation
  is highly influential to the performance: there is a
  sweet-spot, not too big and not too small, between single words and
  full sentences that allows the most meaningful information in a text
  to be effectively retained and recalled. Further, the attention over such
window-based memories can be trained effectively through self-supervision.
  %
  We then  assess the generality of this principle by applying it
  to the CNN QA benchmark, which involves
  identifying named entities in paraphrased summaries of news
  articles, and achieve state-of-the-art performance.
\end{abstract}

\section{Introduction}

Humans do not interpret language in isolation. The context in which words and sentences are understood, whether a conversation, book chapter or road sign, plays an important role in human comprehension \citep{altmann1988interaction,binder2011neurobiology}. In this work, we investigate how well statistical models can exploit such wider contexts to make predictions about natural language.

Our analysis is based on a new benchmark dataset (The Children's Book Test or CBT) designed to test the role of memory and context in language processing and understanding. The test requires predictions about different types of missing words in children's books, given both nearby words and a wider context from the book. Humans taking the test predict all types of word with similar levels of accuracy. However, they rely on the wider context to make accurate predictions about named entities or nouns, whereas it is unimportant when predicting higher-frequency verbs or prepositions. 

As we show, state-of-the-art language modelling architectures, Recurrent Neural Networks (RNNs) with Long-Short Term Memory  (LSTMs), perform differently to humans on this task. They are excellent predictors of prepositions (\emph{on, at}) and verbs (\emph{run, eat}), but lag far behind humans when predicting nouns (\emph{ball, table}) or named entities (\emph{Elvis, France}). This is because their predictions are based almost exclusively on local contexts.  
In contrast, Memory Networks \citep{weston2014memory} are one of a class of `contextual models' that can interpret language at a given point in text conditioned directly on both local information and explicit representation of the wider context. On the CBT, Memory Networks designed in a particular way can exploit this information to achieve markedly better prediction of named-entities and nouns than conventional language models. This is important for applications that require coherent semantic processing and/or language generation, since nouns and entities typically encode much of the important semantic information in language. 


However, not all contextual models reach this level of performance. We find the way in which wider context is represented in memory to be critical.
If memories are encoded from a small window around important words in the context, 
there is an optimal size for memory representations between single words and entire sentences, that depends on the class of word to be predicted. We have nicknamed this effect the \emph{Goldilocks Principle} after the well-known English fairytale~\citep{goldilocks}. 
%
%
In the case of Memory Networks, we also find that self-supervised training of the memory access mechanism yields a clear performance boost when predicting named entities, a class of word that has typically posed problems for neural language models. Indeed, we train a Memory Network with these design features to beat the best reported performance on the CNN QA test of entity prediction from news articles~\citep{nips15_hermann}.

\section{The Children's Book Test}

The experiments in this paper are based on a new resource, the Children's Book Test, designed to measure directly how well language models can exploit wider linguistic context. The CBT is built from books that are freely available thanks to Project Gutenberg.\footnote{\url{https://www.gutenberg.org/}} Using children's books guarantees a clear narrative structure, which can make the role of context more salient. After allocating books to either training, validation or test sets, we formed example `questions' (denoted \(x\)) from chapters in the book by enumerating 21 consecutive sentences. 

\begin{figure}[h]
\centering
\includegraphics[width=\textwidth]{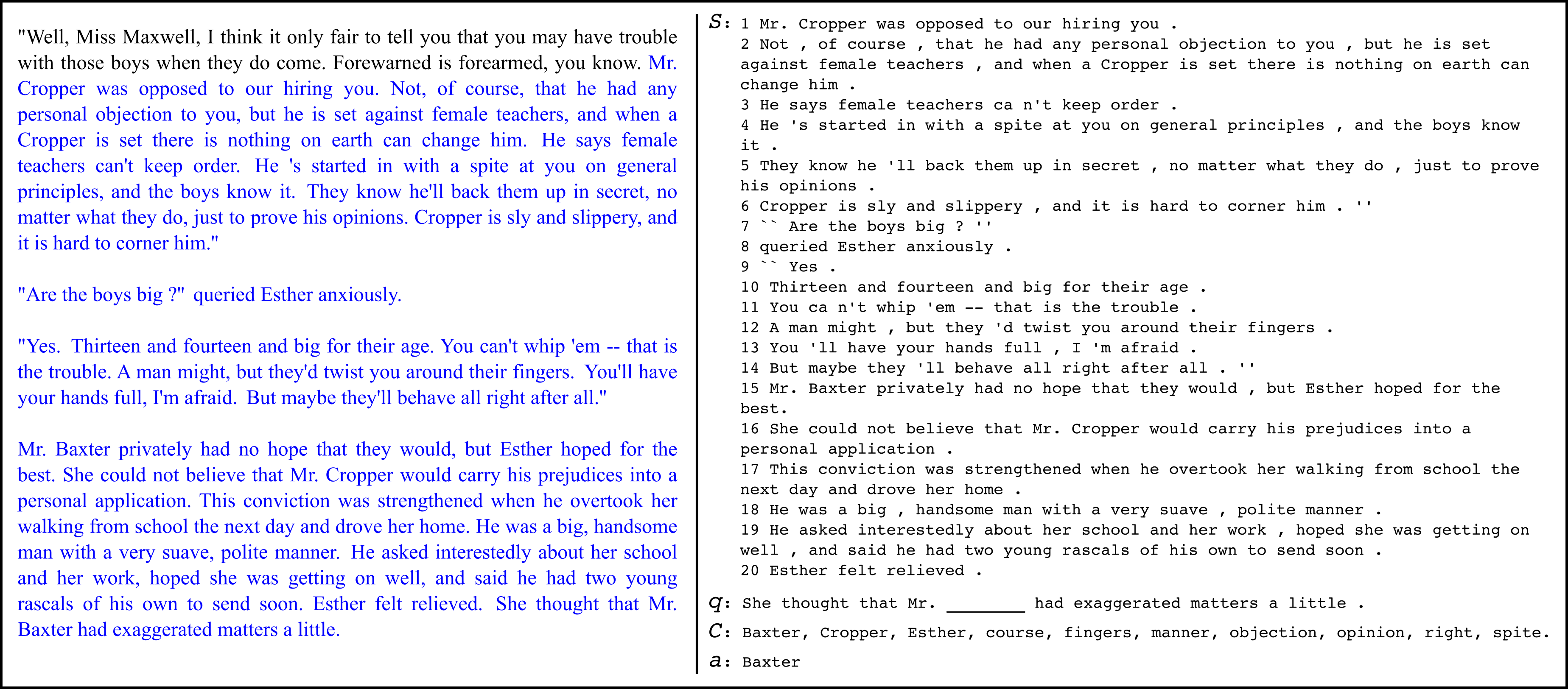}
\caption{{\bf A Named Entity question from the CBT} (right), created from a book passage (left, in blue). In this case, the candidate answers \(C\) are both entities and common nouns, since fewer than ten named entities are found in the context.}
\label{fig:goldilocks}
\end{figure}

In each question, the first 20 sentences form the \emph{context} (denoted \(S\)), and a word (denoted \(a\)) is removed from the 21$^{st}$ sentence, which becomes the \emph{query} (denoted \(q\)). Models must identify the \emph{answer word} \(a\) among a selection of 10 candidate answers (denoted \(C\)) appearing in the context sentences and the query. Thus, for a question answer pair  $(x,\, a)$: $x = (q,\,S,\,C)$; \(S\) is an ordered list of sentences; \(q\) is a sentence (an ordered list \(q=q_1 , \dots q_l\) of words) containing a missing word symbol; \(C\) is  a bag of unique words such that \(a \in C\),  its cardinality \(|C|\) is 10 and every candidate word \(w \in C\) is such that \(w \in q \cup S\). An example question is given in Figure~\ref{fig:goldilocks}.

For finer-grained analyses, we evaluated four classes of question by removing distinct types of word: Named Entities, (Common) Nouns, Verbs and Prepositions (based on output from the POS tagger and named-entity-recogniser in the Stanford Core NLP Toolkit \citep{manning2014stanford}). For a given question class, the nine incorrect candidates are selected at random from words in the context having the same type as the answer. The exact number of questions in the training, validation and test sets is shown in Table \ref{tab:cbt_stat}. Full details of the candidate selection algorithm (e.g. how candidates are selected if there are insufficient words of a given type in the context) can be found with the dataset.\footnote{The dataset can be downloaded from \url{http://fb.ai/babi/}.}

Classical language modelling evaluations are based on average perplexity across all words in a text. They therefore place proportionally more emphasis on accurate prediction of frequent words such as prepositions and articles than the less frequent words that transmit the bulk of the meaning in language \citep{baayen1996word}. In contrast, because the CBT allows focused analyses on semantic content-bearing words, it should be a better proxy for how well a language model can lend semantic coherence to applications including machine translation, dialogue and question-answering systems.

\vspace{-2mm}
\begin{table}[ht]
\label{tab:cbt_stat}
  \begin{center}
    {\small 
      {\sc 
        \begin{tabular}{l|ccc}
          & Training & Validation & Test \\
          \hline
          \hline
          Number of books & 98 & 5 & 5 \\
          Number of questions (context+query)& 669,343 & 8,000 & 10,000  \\
          Average words in contexts & 465 & 435 & 445 \\
          Average words in queries & 31 & 27 & 29 \\
          Distinct candidates & 37,242 & 5,485 & 7,108 \\
          \hline
          Vocabulary size & \multicolumn{3}{|c}{53,628}\\
          \hline
        \end{tabular}
      }
    }
    \caption{\label{tab:cbt_stat} {\bf Statistics of the CBT.} Breakdown by question class is provided with the data set files.}
  \end{center}
\vspace*{-4ex}
\end{table}

\subsection{Related Resources}
There are clear parallels between the CBT and the Microsoft Research Sentence Completion Challenge (MSRCC) \citep{zweig2011microsoft}, which is also based on Project Gutenberg (but not children's books, specifically). A fundamental difference is that, where examples in the MSRCC are made of a single sentence, each query in the CBT comes with a wider context. This tests the sensitivity of language models to semantic coherence beyond sentence boundaries. The CBT is also larger than the MRSCC (10,000 vs 1,040 test questions), requires models to select from more candidates on each question (10 vs 5), covers missing words of different (POS) types and contains large training and validation sets that match the form of the test set. 

There are also similarities between the CBT and the CNN/Daily Mail (CNN QA) dataset recently released by \cite{nips15_hermann}. This task requires models to identify missing entities from bullet-point summaries of online news articles. The CNN QA task therefore focuses more on paraphrasing parts of a text, rather than making inferences and predictions from contexts as in the CBT. It also differs in that all named entities in both questions and articles are anonymised so that models cannot apply knowledge that is not apparent from the article. We do not anonymise entities in the CBT, as we hope to incentivise models that can apply background knowledge and information from immediate and wider contexts to the language understanding problem.\footnote{See Appendix~\ref{ap:anon} for a sense of how anonymisation changes the CBT.} At the same time, the CBT can be used as a benchmark for general-purpose language models whose downstream application is semantically focused generation, prediction or correction. 
The CBT is also similar to the MCTest of machine comprehension \citep{richardson2013mctest}, in which children's stories written by annotators are accompanied by four multiple-choice questions. However, it is very difficult to train statistical models only on MCTest because its training set consists of only 300 examples.

\section{Studying Memory Representation with Memory Networks}

\label{sec:memnn}

Memory Networks \citep{weston2014memory} have shown promising performance at various tasks such as reasoning on the bAbI tasks \citep{weston2015towards} or language modelling \citep{sukhbaatar2015end}. Applying them on the CBT enables us to examine the impact of various ways of encoding  context on their semantic processing ability over naturally occurring language.

\subsection{Encoding Memories and Queries}


Context sentences of $S$ are encoded into memories, denoted $m_i$, using a feature-map \(\phi(s)\) mapping sequences of words \(s \in S\) from the context to one-hot representations in \([0,1]^d\), where $d$ is typically the size of the word vocabulary. We considered several formats for storing the phrases \(s\):
\begin{itemize}[leftmargin=3mm]

\item {\bf Lexical memory:} Each word occupies a separate slot in the memory (each phrase \(s\) is a single word and \(\phi(s)\) has only one non-zero feature). To encode word order, time features are added as embeddings indicating the index of each memory, following \cite{sukhbaatar2015end}. 

\item {\bf Window memory:} Each phrase \(s\) corresponds to a window of text from the context $S$ centred on an individual mention of a candidate $c$ in $S$. Hence, memory slots are filled using windows of words \(\{ w_{i-(b-1)/2} \dots w_i \dots w_{i+(b-1)/2} \} \) where \(w_i\in C\) is an instance of one of the candidate words in the question.\footnote{See Appendix~\ref{ap:nonsparse-windows} for discussion and analysis of using candidates in window representations and training.}
 Note that the number of phrases \(s\) is typically greater than $|C|$ since candidates can occur multiple times in $S$. The window size \(b\) is tuned on the validation set. 
We experimented with encoding as a standard bag-of-words, or
 by having one dictionary per window position, where the latter performed best.
%

\item {\bf Sentential memory:} This setting follows the original implementation of Memory Networks for the bAbI tasks where the phrases \(s\) correspond to complete sentences of $S$.  For the CBT, this means that each question yields exactly 20 memories. We also use Positional Encoding (PE) as introduced by \cite{sukhbaatar2015end} to encode the word positions. 
\end{itemize}

The order of occurrence of memories is less important for sentential and window formats than for lexical memory. So, instead of using a full embedding for each time index, we simply use a scalar value which indicates the position in the passage, ranging from 1 to the number of memories. An additional parameter (tuned on the validation set) scales the importance of this feature. As we show in Appendix~\ref{ap:qa_cnn_ab_study}, time features only gave a marginal performance boost in those cases.


For sentential and window memory formats, queries are encoded in a similar way to the memories: as a bag-of-words representation of the whole sentence and a window of size $b$ centred around the missing word position respectively.
For the lexical memory, memories are made of the $n$ words preceding the word to be predicted, whether these $n$ words come from the context or from the query, and the query embedding is set to a constant vector $0.1$. 

\subsection{End-to-end Memory Networks} \label{sec:mod_o}

The {MemN2N} architecture, introduced by \cite{sukhbaatar2015end}, allows for a direct training of Memory Networks through backpropagation, and consists of two main steps.

First, `supporting memories', those useful to find the correct answer to the query $q$, are retrieved. This is done by embedding both the query and all memories into a single space of dimension $p$ using an embedding matrix $\bA\in\Re^{p\times d}$ yielding the query embedding  $\bq=\bA\phi(q)$ and memory embeddings $\{\bc_i=\bA\phi(s_i)\}_{i=1,\dots n}$, with $n$ the number of memories.
%
The match between $\bq$ and each memory $\bc_i$ in the embedding space is fed through a softmax layer giving a distribution \(\{\alpha_i\}_{i=1, \dots n}\) of matching scores which are used as an \emph{attention} mechanism over the memories to return the first supporting memory:
\begin{equation} \label{eq:eq_o}
  \bm_{o1} = \sum_{i=1 \dots n} \alpha_{i} \bm_i\,, \mbox{~~~~~~}\text{with~~~~~}\alpha_i =  \frac{e^{\bc_i^\top\bq}}{\sum_j e^{\bc_j^\top\bq}}, \, i=1,\dots n,
\end{equation}
and where $\{\bm_i\}_{i=1,\dots n}$ is a set of memory embeddings obtained in the same way as the $\bc_i$, but using another embedding matrix $\bB\in\Re^{p\times d}$.


A characteristic of Memory Networks is their ability to perform several hops in the memory before returning an answer.
Hence the above process can be repeated $K$ times by recursively using $\bq^k = \bq^{k-1} + \bm_{ok-1}$ instead of the original $\bq^1=\bq$. There are several ways of connecting the layers corresponding to distinct hops. We chose to share the embedding matrices $\bA$ and $\bB$ across all layers and add a linear mapping across hops, that is $\bq^k = \bH \bq^{k-1} + \bm_{ok-1}$ with $\bH \in \Re^{p\times p}$. For the lexical memory setting, we also applied ReLU operations to half of the units in each layer following \cite{sukhbaatar2015end}.\footnote{For the lexical memory we use the code available at \url{https://github.com/facebook/MemNN}.}


In a second stage, an answer distribution \(\hat{\ba}=\softmax( \bU \bq^{K+1})\) is returned given $K$ retrieved memories \( \textbf{m}_{o1}, \dots \textbf{m}_{oK} \) and the query \(q\).
Here, $\bU\in\Re^{d\times p}$ is a separate weight matrix that can potentially be tied with $\bA$, and $\hat{\ba}\in\Re^d$ is a distribution over the whole vocabulary. The predicted answer $\hat{a}$ among candidates is then simply $\hat{a}=\argmax_{w\in C}\hat{\ba}(c)$, with $\hat{\ba}(w)$ indicating the probability of word $w$ in $\hat{\ba}$.
For the lexical memory variant, $\hat{a}$ is selected not only by using the probability of each of the ten candidate words, but also of any words that follow the missing word marker in the query.

During training, $\hat{\ba}$  is used to minimise a standard cross-entropy loss with the true label $a$ against all other words in the dictionary (i.e. the candidates are not used in the training loss),  and optimization is carried out using stochastic gradient descent (SGD).
%
Extra experimental details and hyperparameters are given in Appendix~\ref{ap:hp}.


\subsection{Self-supervision for Window Memories} \label{sec:ssup}


After initial experiments, we observed that the capacity to execute multiple hops in accessing memories was only beneficial in the lexical memory model. We therefore also tried a simpler, single-hop Memory Network, i.e. using a single memory to answer, that exploits a stronger signal for learning memory access. A related approach was successfully applied by \cite{bordes2015large} to question answering about knowledge bases.

Memory supervision (knowing which memories to attend to) is not provided at training time but is inferred automatically using the following procedure: since we know the correct answer during training, we hypothesize the correct supporting memory to be among the window memories whose corresponding candidate is the correct answer.		
In the common case where more than one memory contains the correct answer, the model picks the single memory  $\tilde{m}$ that is already scored highest by itself, i.e. scored highest by the query in the embedding space defined by $\bA$.\footnote{TF-IDF similarity worked almost as well in our experiments, but a random choice over positives did not.} 

We train by making gradient steps using SGD to force the model, for each example, to give a higher score to the supporting memory $\tilde{m}$ relative to any other memory from any other candidate.
Instead of using  eq~\eqref{eq:eq_o}, the model selects its top relevant memory using:
\begin{equation} \label{eq:salient}
  m_{o1} = \argmax_{i=1,\dots n} \bc_i^\top\bq \, .
\end{equation}
If $m_{o1}$ happens to be different from $\tilde{m}$, then the model is updated.

At test time, rather than use a hard selection as in eq~\eqref{eq:salient}
the model scores each candidate not only with its highest scoring memory
but with the sum of the scores of all its corresponding windows after passing all scores through a softmax.  That is, the score of a candidate is defined by the sum of the $\alpha_i$ (as used in eq~\eqref{eq:eq_o}) of the windows it appears in. This relaxes the effects of the \(max\) operation and allows for all windows associated with a candidate to contribute some information about that candidate. As shown in the ablation study in Appendix~\ref{ap:qa_cnn_ab_study}, this results in slightly better performance on the CNN QA benchmark compared to hard selection at test time.

Note that self-supervised Memory Networks do not exploit any new label information beyond the training data. 
The approach can be understood as a way of achieving \emph{hard attention} over memories, to contrast with the \emph{soft attention}-style selection described in Section~\ref{sec:mod_o}. Hard attention yields significant improvements in image captioning \citep{xu2015show}. However, where \cite{xu2015show} use the REINFORCE algorithm \citep{williams1992simple} to train through the max of eq~\eqref{eq:salient}, our self-supervision heuristic permits direct backpropagation.




\section{Baseline and Comparison Models}

In addition to memory network variants, we also applied many different types of
 language modelling and machine reading architectures to the CBT.

\subsection{Non-Learning Baselines}
We implemented two simple baselines based on word frequencies. For the first, we selected the most frequent candidate in the entire training corpus. In the second, for a given question we selected the most frequent candidate in its context. In both cases we broke ties with a random choice. 

We also tried two more sophisticated ways to rank the candidates that do not require any learning on the training data. The first is the `sliding window' baseline applied to the MCTest by \cite{richardson2013mctest}. In this method, ten `windows' of the query concatenated with each possible candidate are slid across the context word-by-word, overlapping with a different subsequence at each position. The overlap score at a given position is simply word-overlap weighted TFIDF-style based on frequencies in the context (to emphasize less frequent words). The chosen candidate corresponds to the window that achieves the maximum single overlap score for any position. Ties are broken randomly. 

The second method is the word distance benchmark applied by \cite{nips15_hermann}. For a given instance of a candidate \(w_i\) in the context, the query \(q\) is `superimposed' on the context so that the missing word lines up with \(w_i\), defining a subsequence \(s\) of the context. For each word \(q_i\) in \(q\), an alignment penalty $P = \min( \min_{j = 1 \dots |s|} \{|i - j| : s_j = q_i\}, m)$ is incurred. The model predicts the candidate with the instance in the context that incurs the lowest alignment penalty. We tuned the maximum single penalty \(m=5\)  on the validation data.  

\subsection{N-gram Language Models}
We trained an n-gram language model using the KenLM toolkit \citep{Heafield-estimate}. We used Knesser-Ney smoothing, and a window size of 5, which performed best on the validation set.
We also compare with a variant of language model with cache \citep{kuhn1990cache}, where we linearly interpolate the n-gram model probabilities with unigram probabilities computed on the context.

\subsection{Supervised Embedding Models}
To directly test how much of the CBT can be resolved by good quality dense representations of words (word embeddings), we implement a supervised embedding model similar to that of \citep{weston2010large}. In these models we learn both input and output embedding matrices \(\bA,\,\bB \in \Re^{p\times d}\) for each word in the vocabulary ($p$ is still the embedding dimension and $d$ the vocabulary size). For a given input passage \(q\) and possible answer word \(w\), the score is computed as \(S(q,\,w) = \phi(q) \bA ^\top \bB \phi(w) \), with \(\phi\) the feature function defined in Section~\ref{sec:memnn}.
These models can be considered as lobotomised Memory Networks with zero hops, 
i.e. the attention over the memory component is removed.

We encode various parts of the question as the input passage: the entire {\bf context + query}, just the {\bf query}, a sub-sequence of the query defined by a {\bf window} of maximum \(b\) words centred around the missing word, and a version ({\bf window + position}) in which we use a different embedding matrix  for encoding each position of the window. We tune the window-size \(d=5\) on the validation set.

\subsection{Recurrent Language Models}
We trained probabilistic RNN language models with LSTM activation units on the training stories (5.5M words of text) using minibatch SGD to maximise the negative log-likelihood of the next word. Hyper-parameters were tuned on the validation set. The best model had both hidden layer and word embeddings of dimension $512$.
When answering the questions in the CBT, we allow one variant of this model ({\bf context + query}) to `burn in' by reading the entire context followed by the query and another version to read only the {\bf query} itself (and thus have no access to the context). Unlike the canonical language-modelling task, all models have access to the query words {\em after} the missing word (i.e if $k$ is the position of the missing word, we rank candidate \(c\) based on \(p(q_1 \dots q_{k-1}, c , q_{k+1} \dots q_l)\) rather than simply \(p(q_1 \dots q_{k-1},c)\)).

\cite{mikolov2012context} previously observed performance boosts for recurrent language models by adding the capacity to jointly learn a document-level representation. We similarly apply a context-based recurrent model to our language-modelling tasks, but opt for the convolutional representation of the context applied by \cite{rush2015neural} for summarisation. Our Contextual LSTM (CLSTM) learns a convolutional attention over windows of the context given the objective of predicting all words in the query. We tuned the window size (\(w=5\)) on the validation set. As with the standard LSTM, we trained the CLSTM on the running-text of the CBT training set (rather than the structured query and context format used with the Memory Networks) since this proved much more effective, and we  report results in the best setting for each method.

\subsection{Human Performance}

We recruited 15 native English speakers to attempt a randomly-selected 10\% from each question type of the CBT, in two modes either with question only or with question+context (shown to different annotators), giving 2000 answers in total.
To our knowledge, this is the first time human performance has been quantified on a language modelling task based on 
different word types and context lengths.


\subsection{Other Related Approaches}
The idea of conditioning language models on extra-sentential context is not new. Access to document-level features can improve both classical language models \citep{mikolov2012context} and word embeddings \citep{huang2012improving}. Unlike the present work, these studies did not explore different representation strategies for the wider context or their effect on interpreting and predicting specific word types.

The original Memory Networks \citep{weston2014memory} used hard memory selection with additional labeled supervision for the memory access component, and
were applied to question-answering tasks 
over knowledge bases or simulated worlds. \cite{sukhbaatar2015end} and \cite{kumar2015ask} trained Memory Networks with RNN components end-to-end with soft memory access, and applied them to additional language tasks. The attention-based reading models of \cite{nips15_hermann} also have many commonalities with Memory Networks, differing in word representation choices 
and attention procedures.
Both \cite{kumar2015ask} and \cite{nips15_hermann} propose bidirectional RNNs as a way of representing previously read text. Our experiments in Section~\ref{sec:results} provide a possible explanation for why this is an effective strategy for semantically-focused language processing: bidirectional RNNs naturally focus on small windows of text in similar way to window-based Memory Networks. 

Other recent papers have proposed RNN-like architectures with new ways of reading, storing and updating information to improve their capacity to learn algorithmic or syntactic patterns \citep{joulin2015inferring,dyer2015transition,grefenstette2015learning}. While we do not study these models in the present work, the CBT would be ideally suited for testing this class of model on semantically-focused language modelling.

\section{Results}

\label{sec:results}

\begin{table}[t]
\newcommand{\mc}[1]{\multicolumn{1}{l|}{#1}}
  \begin{center}
    \resizebox{1\linewidth}{!}{
      {\sc 
        \begin{tabular}{l|cccc}
          \mc{Methods} & Named Entities & Common Nouns & Verbs & Prepositions
          \\
          \hline
          \hline
          \mc{Humans (query)$^{(*)}$} & 0.520 & 0.644 & 0.716 & 0.676 \\
          \mc{Humans (context+query)$^{(*)}$} &{\it \textbf{0.816}} & {\it \textbf{ 0.816}} & {\it \textbf{0.828}} & 0.708 \\
          \hline 
          \hline 
          \mc{Maximum frequency (corpus)} & 0.120 & 0.158 & 0.373 & 0.315 \\
          \mc{Maximum frequency (context)} & 0.335 & 0.281 & 0.285 & 0.275 \\
          \mc{Sliding window} & 0.168 & 0.196 & 0.182 & 0.101 \\
          \mc{Word distance model} & 0.398 & 0.364 & 0.380 & 0.237 \\
          \mc{Kneser-Ney language model} & 0.390 & 0.544 & 0.778 & 0.768 \\                                                                    
          \mc{Kneser-Ney language model + cache} & 0.439 & 0.577 & 0.772 & 0.679 \\ 
          \hline
          \mc{\starspace (context+query)} & 0.253 & 0.259 & 0.421 & 0.315 \\
          \mc{\starspace (query)} & 0.351 & 0.400 & 0.614 & 0.535 \\
          \mc{\starspace (window)} & 0.362 & 0.415 & 0.637 & 0.589 \\
          \mc{\starspace (window+position)} & 0.402 & 0.506 & 0.736 & 0.670 \\
          \hline 
          \mc{LSTMs (query)} & 0.408 & 0.541 & 0.813 & 0.802 \\
          \mc{LSTMs (context+query)} & 0.418 & 0.560 & \bf 0.818 & 0.791 \\
          \mc{Contextual LSTMs (window context)} & 0.436 & 0.582 & 0.805 & \bf 0.806\\
          \hline
          \hline
          MemNNs  (lexical memory) &   0.431 & 0.562 & 0.798 & 0.764 \\
          MemNNs  (window memory) &  0.493 & 0.554 & 0.692 & 0.674 \\
          MemNNs  (sentential memory + PE) & 0.318 & 0.305 & 0.502 & 0.326 \\
          MemNNs  (window memory + self-sup.) & \bf 0.666 & \bf 0.630 & 0.690 & 0.703\\
          \hline 
        \end{tabular}
      }
    }
    \caption{\label{tab:cbt_res} {\bf Results on CBT test set.} $^{(*)}$Human results were
      collected on 10\% of the test set.}\label{tab:cbt_res}
  \end{center}
  \vspace*{-4ex}
\end{table}

\paragraph{Modelling syntactic flow} 
In general, there is a clear difference in model performance according
to the type of word to be predicted. Our main results in Table \ref{tab:cbt_res}
show  conventional language models are very good at
predicting prepositions and verbs, but less good at predicting named
entities and nouns. Among these language models, and in keeping with
established results, RNNs with LSTMs demonstrate a small gain on
n-gram models across the board, except for named entities where the cache is beneficial. 
In fact, LSTM models are better than humans at predicting prepositions, which suggests that there are cases in which several of the candidate prepositions are `correct', but annotators prefer the less frequent one.  Even more surprisingly, when only local context (the query) is available, both LSTMs and n-gram models predict verbs more accurately than humans. This may be because the models are better attuned to the distribution of verbs in children's books, whereas humans are unhelpfully influenced by their wider knowledge of all language styles.\footnote{We did not require the human annotators warm up by reading the 98 novels in the training data, but this might have led to a fairer comparison.} When access to the full context is available, humans do predict verbs with slightly greater accuracy than RNNs. 

\paragraph{Capturing semantic coherence} The best performing Memory
Networks predict common nouns and named entities more
accurately than conventional language models. Clearly, in doing so,
these models rely on access to the wider context (the
supervised {\small \sc{embedding model (query)}}, which is equivalent to the
memory network but with no contextual memory, performs poorly in this
regard). The fact that LSTMs without attention perform similarly on
nouns and named entities whether or not the context is available
confirms that they do not effectively exploit this context. This may
be a symptom of the difficulty of storing and retaining information
across large numbers of time steps that has been previously observed
in recurrent networks (See e.g. \cite{bengio1994learning}). 


\paragraph{Getting memory representations `just right'} Not all memory
networks that we trained exploited the context to achieve
decent prediction of nouns and named entities.
For instance, when each sentence in the context is stored as an ordered sequence of word
embeddings (\emph{sentence mem + PE}), performance is quite
poor in general. Encoding the context as an unbroken sequence of individual
words (\emph{lexical memory}) works well for capturing prepositions
and verbs, but is less effective with nouns and entities. In contrast,
\emph{window memories} centred around the
candidate words are more useful than either word-level or
sentence-level memories when predicting named entities and nouns.

%

\begin{figure}[ht]
\newcommand{\mc}[1]{\multicolumn{2}{l}{#1}}
  \begin{center}
   \includegraphics[width=\textwidth]{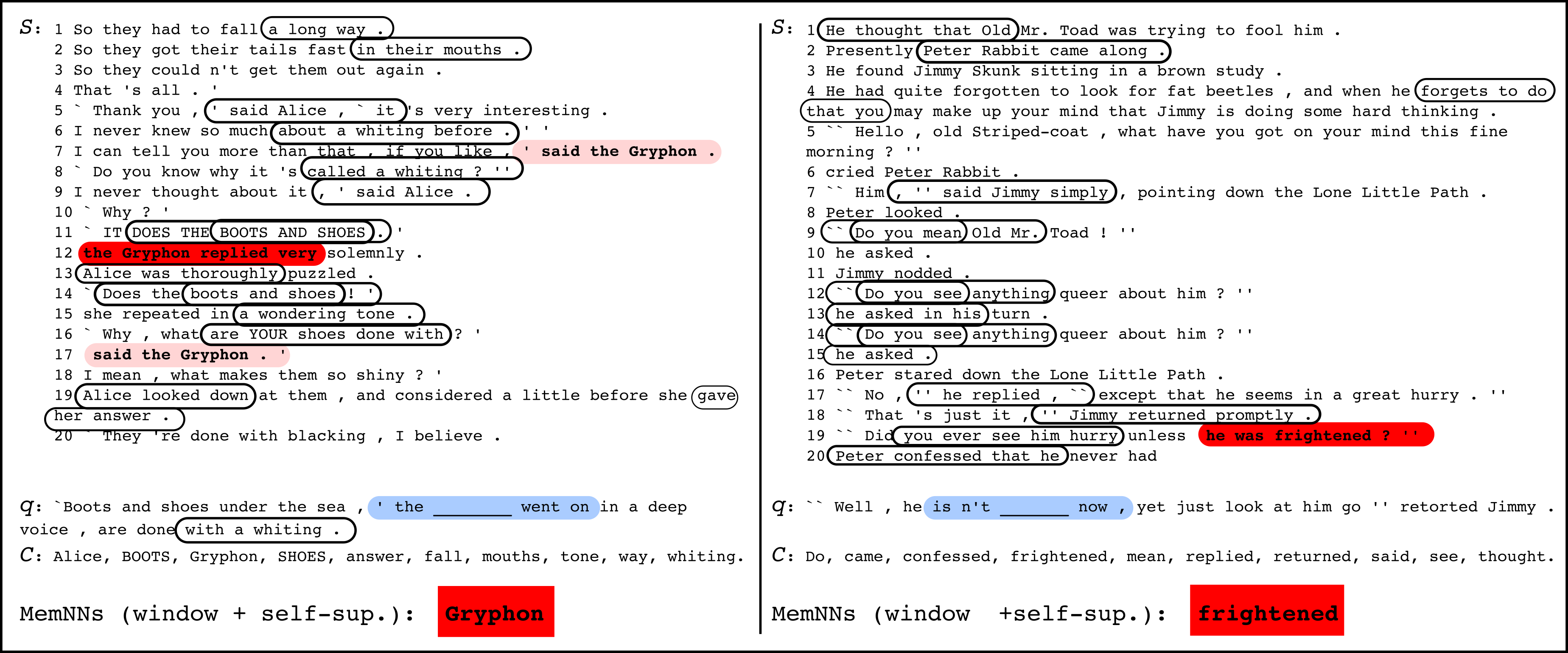}
      \caption{\label{tab:ex_pred_cbt} {\bf Correct predictions of
          MemNNs (window memory + self-supervision) on CBT} on Named Entity (left) and
          Verb (right). Circled phrases indicate all considered
          windows; red ones are the ones corresponding to the returned
          (correct) answer; the blue windows represent the queries.}\label{tab:ex_pred_cbt}
    \end{center}
  \vspace*{-2ex}
\end{figure}

\paragraph{Self-supervised memory retrieval} 
The window-based Memory Network with self-supervision (in which a hard attention selection 
is made among window memories during training)
outperforms all others at predicting named entities and common nouns.
%
%
Examples of predictions made by this model for two CBT questions are shown
in Figure~\ref{tab:ex_pred_cbt}. It is notable that this model is able to achieve the strongest performance with only a simple window-based strategy for representing questions. 

\begin{table}[ht]
  \begin{center}
   \label{tab:qacnn_res}
    \resizebox{1\linewidth}{!}{
      {\sc 
        \begin{tabular}{l|cc}
          Methods & Validation & Test \\
          \hline
          \hline
          Maximum frequency (article)$^{(*)}$ & 0.305 & 0.332 \\
          Sliding window & 0.005 & 0.006 \\
          Word distance model$^{(*)}$ & 0.505 & 0.509 \\
          \hline 
          Deep LSTMs (article+query)$^{(*)}$ & 0.550 & 0.570 \\
          Contextual LSTMs (``Attentive reader'')$^{(*)}$ & 0.616 & 0.630 \\
          Contextual LSTMs (``Impatient reader'')$^{(*)}$ & 0.618 & 0.638 \\
          \hline
          MemNNs (window memory) & 0.580 & 0.606 \\
          MemNNs (window memory + self-sup.) & 0.634 & 0.668 \\
          \hline
          MemNNs (window memory + ensemble) & 0.612 & 0.638 \\ 
          MemNNs (window memory + self-sup.  + ensemble) & 0.649 & 0.684 \\
          \hline
          MemNNs (window  + self-sup.  +  ensemble + exclud. coocurrences) & \bf 0.662 & \bf 0.694 \\
          \hline
        \end{tabular}
      }
    }
    \caption{\label{tab:qacnn_res} {\bf Results on CNN QA.} $^{(*)}$Results taken from \cite{nips15_hermann}.}\label{cnn}
  \end{center}
  \vspace*{-2ex}
\end{table}

\subsection{News Article Question Answering}
To examine how well our conclusions generalise to different machine
reading tasks and language styles, we also tested the
best-performing Memory Networks on the CNN QA task \citep{nips15_hermann}.\footnote{The CNN QA dataset was released after our primary experiments were completed, hence we experiment only with one of the two large datasets released with that paper.} This dataset consists of 93k news articles from the CNN website, each coupled with a question derived from a bullet point summary accompanying the article, and a single-word answer. The answer is always a named entity, and all named entities in the article function as possible candidate answers.

As shown in Table~\ref{cnn}, our window model without self-supervision 
achieves similar
performance to the best approach proposed for the task by \cite{nips15_hermann}
when using an ensemble of MemNN models.
Our use of an ensemble 
is an alternative way of replicating the application of \emph{dropout}~\citep{hinton2012improving} in the previous best
approaches \citep{nips15_hermann} as ensemble averaging has similar effects to dropout~\citep{wan2013regularization}.
When self-supervision is added, the Memory Network
greatly surpasses the state-of-the-art on this task. 
Finally, the last line of
Table~\ref{cnn} (\emph{excluding co-occurrences}) shows how an additional heuristic, removing from the candidate list all named entities
already appearing in the bullet point summary, boosts performance even further.

Some common principles may explain the strong
performance of the best performing models on this
task. The attentive/impatient reading models encode the articles using bidirectional RNNs
\citep{graves2008unconstrained}. For each word in
the article, the combined hidden state of such an RNN naturally
focuses on a window-like chunk of surrounding text, much like the window-based memory network or the CLSTM. Together, these results therefore support the principle that the most informative representations of text correspond to sub-sentential chunks. Indeed, the observation that the most informative representations
for neural language models correspond to small chunks of text is
also consistent with recent work on neural machine translation, in which
\cite{luong2015effective} demonstrated improved performance by
restricting their attention mechanism to small  windows of
the source sentence.

Given these commonalities in how the reading models and Memory
Networks represent context, the advantage of the best-performing
Memory Network instead seems to stem from how it accesses or retrieves
this information; in particular, the hard attention and
self-supervision. Jointly learning to access and use information is a
difficult optimization. Self-supervision in particular makes effective
Memory Network learning more tractable.\footnote{See the appendix for an ablation study in which optional features of the memory network are removed.}

\section{Conclusion}
We have presented the Children's Book Test, a new semantic language modelling benchmark. The CBT measures how well models can use both local and wider contextual information to make predictions about different types of words in children's stories. By separating the prediction of syntactic function words from more semantically informative terms, the CBT provides a robust proxy for how much language models can impact applications requiring a focus on semantic coherence. 

We tested a wide range of models on the CBT, each with different ways of
representing and retaining previously seen content. This enabled us to draw novel insights into the optimal strategies for representing and accessing semantic information in memory. One consistent finding was that memories that encode sub-sentential chunks (windows) of informative text 
seem to be most useful to neural nets when interpreting and modelling language. 
However, our results indicate that the
most useful text chunk size depends on the modeling task 
(e.g. semantic content vs. syntactic function words).
%
We showed that Memory Networks that adhere to this principle can be efficiently trained using a simple
self-supervision to surpass all other methods for predicting named
entities on both the CBT and the CNN QA
benchmark, an independent test of machine reading. 
%

\subsubsection*{Acknowledgments}
The authors would like to thank Harsha Pentapelli and Manohar Paluri
for helping to collect the human annotations and Gabriel Synnaeve for
processing the QA CNN data.

\bibliography{iclr2016_conference}
\bibliographystyle{iclr2016_conference}

\appendix

\section{Experimental Details} \label{ap:hp}

\paragraph{Setting}
The text of questions is lowercased for all Memory Networks as
well as for all non-learning baselines. LSTMs models use the raw text (although we also tried lowercasing, which made little difference).
Hyperparameters of all learning models have been set using
grid search on the validation set.
The main hyperparameters are embedding dimension $p$, learning rate
$\lambda$, window size $b$, number of hops $K$, maximum memory size
$n$ ($n=all$ means using all potential memories).
All models were implemented using the Torch library (see {\tt torch.ch}).
For CBT, all models have been trained on all question types 
altogether. 
We did not try to experiment with word embeddings pre-trained on a
bigger corpus.

\paragraph{Optimal hyper-parameter values on CBT:}

\begin{itemize}
\item Embedding model (context+query): $p=300$, $\lambda=0.01$.
\item Embedding model (query): $p=300$, $\lambda=0.01$.
\item Embedding model (window): $p=300$, $\lambda=0.005$, $b=5$.
\item Embedding model (window+position): $p=300$, $\lambda=0.01$, $b=5$.
\item LSTMs (query \& context+query): $p=512$, $\lambda=0.5$, $1$
  layer, gradient clipping factor: $5$, learning rate shrinking factor: $2$.
\item Contextual LSTMs: $p=256$, $\lambda=0.5$, $1$
  layer, gradient clipping factor: $10$, learning rate shrinking
  factor: $2$.
\item MemNNs  (lexical memory): $n=200$, $\lambda=0.01$, $p=200$, $K=7$.
\item MemNNs  (window  memory): $n=all$, $b=5$, $\lambda=0.005$,
  $p=100$, $K=1$.
\item  MemNNs  (sentential memory + PE): $n=all$, $\lambda=0.001$,
  $p=100$, $K=1$.
\item MemNNs  (window  memory + self-sup.): $n=all$, $b=5$, $\lambda=0.01$,
  $p=300$.
\end{itemize}

\paragraph{Optimal hyper-parameter values on CNN QA:}

\begin{itemize}
\item  MemNNs (window memory): $n=all$, $b=5$, $\lambda=0.005$,
  $p=100$, $K=1$.
\item  MemNNs (window memory + self-sup.):  $n=all$, $b=5$, $\lambda=0.025$,
  $p=300$, $K=1$.
\item  MemNNs (window memory + ensemble): $7$ models with $b=5$.
\item  MemNNs (window memory + self-sup.  + ensemble): $11$ models with $b=5$.
\end{itemize}

\section{Results on CBT Validation Set} \label{ap:res_val}

\newcommand{\mc}[1]{\multicolumn{1}{l|}{#1}}
  \begin{center}
  \vspace*{-2ex}
    \resizebox{1\linewidth}{!}{
      {\sc 
        \begin{tabular}{l|cccc}
          \mc{Methods} & Named Entities & Common Nouns & Verbs & Prepositions
          \\
          \hline 
          \hline 
          \mc{Maximum frequency (corpus)} & 0.052 & 0.192 & 0.301 & 0.346 \\
          \mc{Maximum frequency (context)} & 0.299 & 0.273 & 0.219 & 0.312 \\
          \mc{Sliding window} & 0.178 & 0.199 & 0.200 & 0.091 \\
          \mc{Word distance model} & 0.436 & 0.371 & 0.332 & 0.259 \\
          \mc{Kneser-Ney language model} & 0.481 & 0.577 & 0.762 & 0.791 \\ 
          \mc{Kneser-Ney language model + cache} & 0.500 & 0.612 & 0.755 & 0.693 \\ 
          \hline
          \mc{\starspace (context+query)} & 0.235 & 0.297 & 0.368 & 0.356 \\
          \mc{\starspace (query)} & 0.418 & 0.462 & 0.575 & 0.560 \\
          \mc{\starspace (window)} & 0.457 & 0.486 & 0.622 & 0.619 \\
          \mc{\starspace (window+position)} & 0.488 & 0.555 & 0.722 & 0.683 \\
          \hline 
          \mc{LSTMs (query)} & 0.500 & 0.613 & 0.811 & \bf 0.819 \\
          \mc{LSTMs (context+query)} & 0.512 & 0.626 & \bf 0.820 & 0.812 \\
          \mc{Contextual LSTMs (window context)} & 0.535 & 0.628 & 0.803 & 0.798 \\
          \hline
          \hline
          MemNNs  (lexical memory) &   0.519 & 0.647 & 0.818 & 0.785 \\
          MemNNs  (window  memory) &  0.542 & 0.591 & 0.693 & 0.704 \\
          MemNNs  (sentential memory + PE) & 0.297 & 0.342 & 0.451 &
                                                                     0.360 \\
          \hline
          MemNNs  (window  memory + self-sup.) & \bf 0.704 & \bf 0.642 & 0.688 & 0.696\\
          \hline
        \end{tabular}
      }
    }
  \end{center}

\section{Ablation Study on CNN QA} \label{ap:qa_cnn_ab_study}
  \begin{center}
    \vspace*{-4ex}
    \resizebox{1\linewidth}{!}{
      {\sc 
        \begin{tabular}{l|cc}
          Methods & Validation & Test \\
          \hline
          \hline
          MemNNs (window memory + self-sup. +  exclud. coocurrences)  &  0.635 & 0.684 \\
          MemNNs (window memory + self-sup.)    &  0.634 & 0.668 \\
          MemNNs (window mem. + self-sup.) -time     & 0.625 & 0.659 \\
          MemNNs (window mem. + self-sup.) -soft memory weighting         & 0.604 & 0.620 \\
          MemNNs (window mem. + self-sup.) -time -soft memory weighting  & 0.592 & 0.613 \\
          \hline
          MemNNs (window mem. + self-sup. + ensemble)              & 0.649 & 0.684 \\
          MemNNs (window mem. + self-sup. + ensemble) -time        &  0.642 &  0.679 \\
          MemNNs (window mem. + self-sup. + ensemble) -soft memory weighting & 0.612  &  0.641 \\
          MemNNs (window mem. + self-sup. + ensemble) -time -soft memory weighting &  0.600 &  0.640 \\
          \hline
        \end{tabular}
      }
    }
({\it Soft memory weighting}: the softmax to select
the best candidate in test as defined in Section~\ref{sec:ssup})
  \end{center}

\section{Effects of Anonymising Entities in CBT} \label{ap:anon}
  \begin{center}
    \vspace*{-4ex}
    \resizebox{1\linewidth}{!}{
      {\sc 
        \begin{tabular}{l|cccc}
          Methods & Named Entities & Common Nouns & Verbs & Prepositions
          \\
          \hline
          MemNNs (word mem.) & 0.431 & 0.562 & 0.798 & 0.764 \\
          MemNNs (\window mem.) & 0.493 & 0.554 & 0.692 & 0.674 \\
          MemNNs (sentence mem.+PE) & 0.318 & 0.305 & 0.502 & 0.326 \\
          MemNNs (\window mem.+self-sup.) &  0.666 &  0.630 & 0.690 & 0.703\\
          \hline
          ANONYMIZED MemNNs (\window +self-sup.) & 0.581 & 0.473 & 0.474 & 0.522\\
          \hline
        \end{tabular}
      }
    }
  \end{center}

  To see the impact of the anonymisation of entities and words as done
  in CNN QA on the self-supervised Memory Networks on the CBT, we
  conducted an experiment where we replaced the mentions of the ten
  candidates in each question by anonymised placeholders in train,
  validation and test. The table above shows results on CBT test set
  in an anonymised setting (last row) compared to MemNNs in a
  non-anonymised setting (rows 2-5).  Results indicate that this has a
  relatively low impact on named entities but a larger one on more
  syntactic tasks like prepositions or verbs.


\section{Candidates and Window Memories in CBT} \label{ap:nonsparse-windows}

In our main results in Table \ref{tab:cbt_res} the window memory is constructed as the set of windows
over the candidates being considered for a given question.
Training of {\sc MemNNs (window memory)} is performed by making gradient steps for questions, with 
the true answer word as the target compared against all words in the dictionary as described in Sec. \ref{sec:mod_o}.
Training of {\sc MemNNs  (window memory + self-sup.)} is performed by making gradient steps for questions, 
with the true answer word as the target compared against all other candidates as described in Sec. \ref{sec:ssup}.
As {\sc MemNNs  (window memory + self-sup.)} is the best performing method for named entities and common nouns, 
to see the impact of these choices we conducted some further experiments with variants of it.

Firstly, window memories do not have to be restricted to candidates, we could consider all possible windows.
Note that this does not make any difference at evaluation time on CBT as one would still evaluate by multiple choice using the candidates, and those extra windows would not contribute to the scores of the candidates.
However, this may make a difference to the weights if used at training time.
We call this ``all windows'' in the experiments to follow.

Secondly, the self-supervision process does not have to rely on there being known candidates:
all that is required is a positive label, in that case we can perform gradient steps with
the true answer word as the target compared against all words in the dictionary (as opposed to only candidates)
as described in Sec. \ref{sec:mod_o}, 
while still using hard attention supervision as described in \ref{sec:ssup}. 
We call this ``all targets'' in the experiments to follow.

Thirdly, one does not have to try to train on only the {\em questions} in CBT, but can treat the children's 
books as a standard 
language modeling task. In that case, {\em all targets} and {\em all windows} must be used, as multiple choice questions have not been constructed for every single word (although indeed many of them are covered by the four word classes). We call
this ``LM'' (for language modeling) in the experiments to follow.

Results with these alternatives are presented in Table \ref{ap:windoze}, the new variants are the last three rows.
Overall, the differing approaches have relatively little impact on the results, as all of them provide superior
results on named entities and common nouns than without self-supervision.
However, we note that 
the use of all windows or LM rather than candidate windows does impact training and testing speed.

\begin{table}[h]
  \begin{center}
    \resizebox{1\linewidth}{!}{
      {\sc 
        \begin{tabular}{l|cccc}
          \mc{Methods} & Named Entities & Common Nouns & Verbs & Prepositions
          \\
          MemNNs  (lexical memory) &   0.431 & 0.562 & 0.798 & 0.764 \\
          MemNNs  (window memory) &  0.493 & 0.554 & 0.692 & 0.674 \\
          MemNNs  (sentential memory + PE) & 0.318 & 0.305 & 0.502 & 0.326 \\
          MemNNs  (window memory + self-sup.) & \bf 0.666 & \bf 0.630 & 0.690 & 0.703\\
          \hline 
          \hline
          MemNNs  (all windows + self-sup.)                  & 0.648 & 0.604 & 0.711 & 0.693\\
          MemNNs  (all windows + all targets + self-sup.)    & 0.639 & 0.602 & 0.698 & 0.667\\
          MemNNs  (LM + self-sup.)                           & 0.638 & 0.605 & 0.692 & 0.647\\
        \end{tabular}
      }
    }
    \caption{\label{ap:windoze} {\bf Results on CBT test set when considering all windows or targets.}}
  \end{center}
  \vspace*{-4ex}
\end{table}

\end{document}